\documentclass[sigconf]{acmart}

\usepackage[utf8]{inputenc} 
\usepackage[T1]{fontenc}    
\usepackage{hyperref}       
\usepackage{url}            
\usepackage{booktabs}       
\usepackage{nicefrac}       
\usepackage{microtype}      
\usepackage{comment}        
\usepackage{amsmath}
\usepackage{multirow}
\usepackage{bm}
\usepackage{amsthm}
\usepackage{hhline}
\usepackage{makecell}
\usepackage{mathtools}
\usepackage{color}
\usepackage{xcolor}
\usepackage{makecell}
\usepackage{graphicx}
\usepackage{arydshln}
\usepackage{framed}
\usepackage[font=small]{caption}
\usepackage[scientific-notation=true]{siunitx}
\usepackage{wrapfig}
\usepackage{algorithm}
\usepackage{algorithmic}
\usepackage{lipsum}
\usepackage{sidecap}
\usepackage{pifont}
\usepackage{fixltx2e}
\usepackage[flushleft]{threeparttable}
\usepackage{diagbox}
\usepackage{listings}
\newcommand{\cmark}{\ding{51}}%
\newcommand{\xmark}{\ding{55}}%

\newcommand{\namel}{\textsc{DIsentangled Multimodal Explanations}}
\newcommand{\names}{\textsc{DIME}}

\definecolor{gg}{RGB}{15,150,15}
\definecolor{rr}{RGB}{230,45,45}

\def\eqref#1{eq~(\ref{#1})}

\def\1{\bm{1}}

\DeclareMathAlphabet{\mathsfit}{\encodingdefault}{\sfdefault}{m}{sl}
\SetMathAlphabet{\mathsfit}{bold}{\encodingdefault}{\sfdefault}{bx}{n}

\usepackage{hyperref}

\usepackage{booktabs}       %
\usepackage{amsfonts}       %
\usepackage{nicefrac}       %
\usepackage{microtype}      %
\usepackage{subcaption}

\usepackage{enumitem}
\setlist{nolistsep}
\setlist[itemize]{noitemsep, topsep=0pt}

\usepackage{times}

\usepackage{graphicx} %
\usepackage{color}

\usepackage{array}
\usepackage{amsmath}
\usepackage{xspace}
\usepackage{fancyhdr}
\usepackage{comment}

\newcolumntype{H}{>{\setbox0=\hbox\bgroup}c<{\egroup}@{}}

\newcommand{\noaistats}[1]{}  %

\definecolor{darkgreen}{rgb}{0,0.4,0.0}
\definecolor{darkblue}{rgb}{0,0.1,0.3}
\definecolor{darkred}{rgb}{0.7,0.0,0.0}

\AtBeginDocument{%
  \providecommand\BibTeX{{%
    \normalfont B\kern-0.5em{\scshape i\kern-0.25em b}\kern-0.8em\TeX}}}

\author{Yiwei Lyu}
\email{ylyu1@andrew.cmu.edu}
\affiliation{%
  \institution{Carnegie Mellon University}
  \city{Pittsburgh}
  \state{PA}
  \country{USA}
}

\author{Paul Pu Liang}
\email{pliang@cs.cmu.edu}
\affiliation{%
  \institution{Carnegie Mellon University}
  \city{Pittsburgh}
  \state{PA}
  \country{USA}
}

\author{Zihao Deng}
\email{zihaoden@andrew.cmu.edu}
\affiliation{%
  \institution{Carnegie Mellon University}
  \city{Pittsburgh}
  \state{PA}
  \country{USA}
}

\author{Ruslan Salakhutdinov}
\email{rsalakhu@cs.cmu.edu}
\affiliation{%
  \institution{Carnegie Mellon University}
  \city{Pittsburgh}
  \state{PA}
  \country{USA}
}

\author{Louis-Philippe Morency}
\email{morency@cs.cmu.edu}
\affiliation{%
  \institution{Carnegie Mellon University}
  \city{Pittsburgh}
  \state{PA}
  \country{USA}
}

\begin{document}

\title{\names: Fine-grained Interpretations of Multimodal Models via Disentangled Local Explanations}


\renewcommand{\shortauthors}{}

\begin{abstract}
    The ability for a human to understand an Artificial Intelligence (AI) model's decision-making process is critical in enabling stakeholders to visualize model behavior, perform model debugging, promote trust in AI models, and assist in collaborative human-AI decision-making. As a result, the research fields of interpretable and explainable AI have gained traction within AI communities as well as interdisciplinary scientists seeking to apply AI in their subject areas. In this paper, we focus on advancing the state-of-the-art in interpreting multimodal models - a class of machine learning methods that tackle core challenges in representing and capturing interactions between heterogeneous data sources such as images, text, audio, and time-series data. Multimodal models have proliferated numerous real-world applications across healthcare, robotics, multimedia, affective computing, and human-computer interaction. By performing model disentanglement into unimodal contributions (\textbf{UC}) and multimodal interactions (\textbf{MI}), our proposed approach, \names, enables accurate and fine-grained analysis of multimodal models while maintaining generality across arbitrary modalities, model architectures, and tasks. Through a comprehensive suite of experiments on both synthetic and real-world multimodal tasks, we show that \names\ generates accurate disentangled explanations, helps users of multimodal models gain a deeper understanding of model behavior, and presents a step towards debugging and improving these models for real-world deployment. Code for our experiments can be found at \url{https://github.com/lvyiwei1/DIME}.
\end{abstract}


\begin{teaserfigure}
\vspace{-2mm}
  \includegraphics[width=\textwidth]{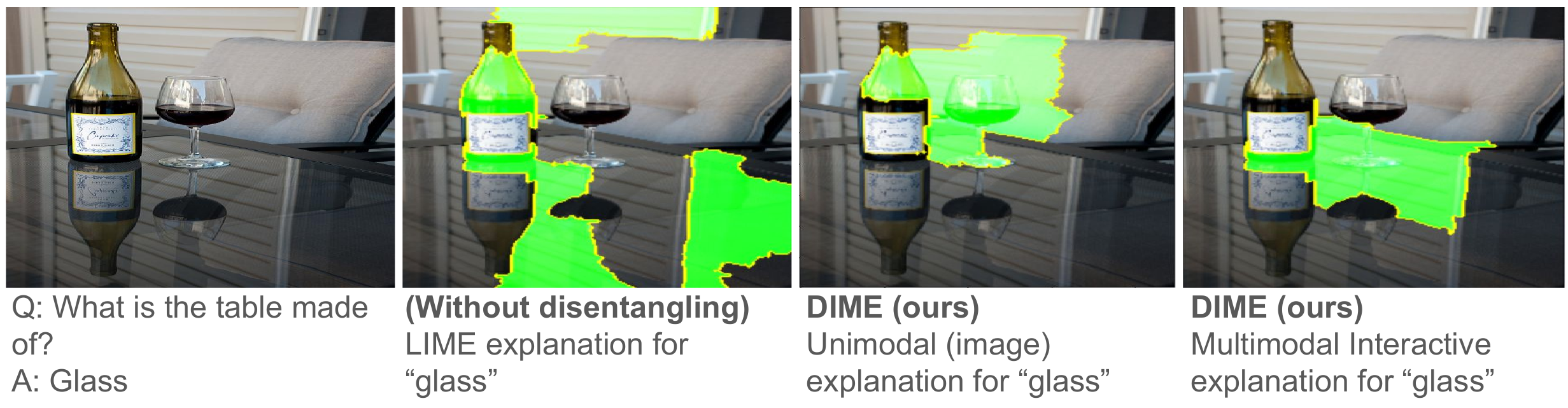}
  \caption{
  \names\ is a novel method of interpreting multimodal models by disentangling the model into unimodal contributions (\textbf{UC}) and multimodal interactions (\textbf{MI}), before generating visual explanations for each. Here is an example from visualizing a trained LXMERT model on VQA: without disentangling, the explanation highlights both parts of the table and the glass bottle; with disentangling, we can see that the unimodal contributions look at the image without looking at the question and highlights the glass bottle, wine glass, and glass table, all of which could support the answer glass, while the multimodal interaction knows that only the table matters, so it focuses only on the table. Therefore, with \names, we can be certain that the model identifies the table as the reason for answering ``glass'', and not the bottle instead. In this paper, we show that \names\ can accurately perform disentanglement and generate explanations for both UC and MI to help researchers better interpret multimodal models.}
  \label{intro}
\end{teaserfigure}

\maketitle

\section{Introduction}

\begin{table*}[]
\centering
\vspace{2mm}
\begin{tabular}{l|cccc}
    \Xhline{3\arrayrulewidth}
    Advantages & \textbf{\names\ (ours)} & LIME~\cite{lime} & EMAP~\cite{hessel2020emap} & M2Lens~\cite{wang2021m2lens} \\
    \hline
    Works for black-box models & \textcolor{gg}\cmark & \textcolor{gg}\cmark & \textcolor{gg}\cmark & \textcolor{gg}\cmark \\
    \hline
    \begin{tabular}[c]{@{}l@{}}Works for modalities from arbitrary\\ classification tasks (not restricted to\\ specific tasks or domains)\end{tabular} & \textcolor{gg}\cmark & \textcolor{gg}\cmark & \textcolor{gg}\cmark & \textcolor{rr}\xmark \\
    \hline
    Visualizes important input features & \textcolor{gg}\cmark & \textcolor{gg}\cmark & \textcolor{rr}\xmark & \textcolor{gg}\cmark \\
    \hline
    \begin{tabular}[c]{@{}l@{}}Disentangles unimodal contributions \\ (UC) and multimodal interactions (MI)\end{tabular} & \begin{tabular}[c]{@{}c@{}} \textcolor{gg}\cmark \\ (RQ1, \S\ref{rq1})\end{tabular} & \textcolor{rr}\xmark & UC only & \textcolor{rr}\xmark \\
    \hline
    \begin{tabular}[c]{@{}l@{}}Determines whether UC or MI (or both)\\ is the dominant factor behind the\\ model's predictions\end{tabular} & \begin{tabular}[c]{@{}c@{}} \textcolor{gg}\cmark \\ (RQ2, \S\ref{rq2})\end{tabular} & \textcolor{rr}\xmark & Sometimes (*) & \textcolor{gg}\cmark \\
    \hline
    \begin{tabular}[c]{@{}l@{}}Provides insight into what features are \\ being aligned or recognized in MI\end{tabular} & \begin{tabular}[c]{@{}c@{}} \textcolor{gg}\cmark \\ (RQ3, \S\ref{rq3})\end{tabular} & \textcolor{rr}\xmark & \textcolor{rr}\xmark & \textcolor{rr}\xmark \\
    \hline
    \begin{tabular}[c]{@{}l@{}}Visualizes each of UC and MI to reveal\\ undesirable model behavior\end{tabular} & \begin{tabular}[c]{@{}c@{}} \textcolor{gg}\cmark \\ (RQ3, \S\ref{rq3})\end{tabular} & \textcolor{rr}\xmark & \textcolor{rr}\xmark & \textcolor{rr}\xmark \\
    \Xhline{3\arrayrulewidth}
\end{tabular}
\vspace{2mm}
\caption{In comparison with related work~\cite{lime,hessel2020emap,wang2021m2lens}, \names\ disentangles a multimodal model into unimodal contributions and multimodal interactions and generates accurate explanations for each, while remaining generalizable (not designed specifically for any model, modality, or task) and works for black-box models (does not require knowledge of the internal structure of the model). \names\ can help human users determine the dominant factor behind a model's decisions, gain insight into what specific multimodal interactions are captured, and reveal undesirable behavior for debugging or improving the model. (*) EMAP can only sometimes distinguish the dominant factor between UC or MI: according to~\citet{hessel2020emap}, EMAP can give insight on individual data points only under special conditions (such as when EMAP happens to flip the prediction).}
\label{tab:overview}
\vspace{-4mm}
\end{table*}

As machine learning models are increasingly deployed in real-world scenarios, it has motivated the development of interpretable machine learning (ML) as a research field with the goal of understanding ML models, performing model debugging, and using these insights to better inform the interaction between AI and humans in joint decision making~\cite{gilpin2018explaining,bhatt2020explainable,chen2022interpretable}. Recently, the promise of multimodal models for real-world representation learning in numerous applications such as multimedia~\cite{liang2021multibench,liang2018multimodal,1667983}, affective computing~\cite{liang2018ranking,PORIA201798}, robotics~\cite{kirchner2019embedded,lee2019making}, finance~\cite{doi:10.1177/0170840618765019}, dialogue~\cite{Pittermann2010}, human-computer interaction~\cite{dumas2009multimodal,obrenovic2004modeling}, and healthcare~\cite{xu2019multimodal} has invigorated research into multimodal machine learning, which brings unique challenges for both computational and theoretical research given the heterogeneity of various data sources and difficulty of capturing correspondences between modalities~\cite{baltruvsaitis2018multimodal}. Among one of these core challenges is \textit{interpretable multimodal learning} with the end goal of empowering various stakeholders by providing insights into multimodal learning, improving model design, or debugging models and datasets.

Recent work in interpretable multimodal learning has therefore focused on constructing interpretable multimodal models via careful model design~\cite{tsai2020multimodal,zadeh2018multimodal,park2018multimodal} or performing post-hoc explanations of black-box multimodal models~\cite{goyal2016towards,chandrasekaran2018explanations}. However, existing works typically focus on building interpretable models using suitable inductive biases, such as designing multimodal routing networks~\cite{tsai2020multimodal}, graph-based fusion~\cite{zadeh2018multimodal}, or multimodal explanation networks to highlight visual importance~\cite{park2018multimodal}. Some of these approaches also require the collection of specialized datasets annotated for visual explanations as intermediate steps in training interpretable models~\cite{park2018multimodal}. On the other hand, with the trend towards large-scale modeling or pre-training as an alternative over individual modality-specific or task-specific models~\cite{visualbert,liang2021multibench}, it is increasingly important to design general-purpose approaches that (1) are able to generate post-hoc explanations for arbitrary black-box models, and (2) does not assume anything about the modality or classification task itself.

As a step towards more fine-grained interpretations of general-purpose multimodal models across arbitrary tasks, we propose \names, an interpretation method for black-box multimodal models. While existing work has been able to generate useful explanations to help humans understand model decision-making processes~\cite{chandrasekaran2018explanations}, they are often only performed at one step of the entire multimodal decision-making process. These singular steps typically include attributing feature importance~\cite{park2018multimodal,chandrasekaran2018explanations} or representation importance~\cite{tsai2020multimodal,zadeh2018multimodal}. The core idea in \names\ is to provide more fine-grained interpretations by disentangling a multimodal model into unimodal contributions (\textbf{UC}) and multimodal interactions (\textbf{MI}). We show that this key insight enables more accurate and fine-grained analysis of multimodal models while maintaining generality across arbitrary modalities, model architectures~\cite{kamath2021mdetr,lxmert}, and tasks~\cite{balanced_vqa_v2,johnson2017clevr}.

Through a comprehensive suite of experiments on both synthetic and real-world multimodal tasks, we show that \names\ is able to accurately perform disentanglement and generate reliable explanations for both UC and MI. Using \names, we are able to gain a deeper understanding of model behavior on challenging multimodal tasks. For example, on VQA 2.0~\cite{goyal2017making}, we successfully use \names\ to determine whether the model uses correct multimodal interactions to answer the questions, as shown in Figure~\ref{intro}. By providing these model explanations to a human annotator, they are able to gain additional insights on model behavior and better determine whether UC, MI, or both are the dominant factor behind the model's predictions on individual datapoints. Furthermore, \names\ presents a step towards debugging and improving these models by systematically revealing certain undesirable behaviors.

\section{Related Work}

Interpretable machine learning as a research field aims to further our understanding of AI models, empower various stakeholders to build trust in AI models, perform model debugging, and use these insights to better inform the interaction between AI and humans in joint decision making~\cite{gilpin2018explaining,bhatt2020explainable,chen2022interpretable}. We cover related concepts in interpreting unimodal models and multimodal models.

\subsection{Interpreting Unimodal Models}

Related work has studied approaches for better understanding unimodal models used for vision, language, and audio modalities. These approaches can be roughly categorized into interpretable ML as designing models which are understandable by design, and explainable ML which focuses on producing post-hoc explanations for black-box models~\cite{rudin2019stop}. In the former, methods such as Concept Bottleneck Models~\cite{koh2020concept} and fitting sparse linear layers~\cite{wong2021leveraging} or decision trees on top of deep feature representations~\cite{wan2020nbdt} have emerged as promising choices marrying the expressive power of deep features with the interpretable decision-making processes of linear models or decision trees. In the latter, approaches such as saliency maps~\cite{simonyan2013deep,smilkov2017smoothgrad}, using surrogate models to interpret local decision boundaries~\cite{lime}, feature visualizations~\cite{yosinski2015understanding,erhan2009visualizing}, and assigning semantic concepts~\cite{bau2017network} all aim to provide insight into model predictions for specific input instances. We refer the reader to~\citet{chen2022interpretable} for a survey and taxonomy of interpretable ML approaches, as well as~\citet{bhatt2020explainable} for an analysis of how interpretable and explainable ML tools can be used in the real world.

\subsection{Interpreting Multimodal Models}

Similar to the interpretation of unimodal models, recent work in interpretable multimodal learning can be categorized into two sections: (1) constructing interpretable multimodal models via careful model design~\cite{tsai2020multimodal,zadeh2018multimodal,park2018multimodal} or (2) performing post-hoc explanations of black-box multimodal models~\cite{goyal2016towards,chandrasekaran2018explanations}. In the former, multimodal routing networks~\cite{tsai2020multimodal}, graph-based fusion techniques~\cite{zadeh2018multimodal,liang2018computational}, multimodal explanation networks to highlight visual importance~\cite{park2018multimodal}, hard-attention~\cite{chen2017multimodal}, and neuro-symbolic reasoning methods~\cite{vedantam2019probabilistic,andreas2016neural} have emerged as strong design choices as a step towards more interpretable multimodal learning. These approaches individually focus on building interpretable components for either modality importance~\cite{park2018multimodal}, cross-modal interactions~\cite{tsai2020multimodal,zadeh2018multimodal,liang2018computational}, or the reasoning process on top of cross-modal interactions~\cite{vedantam2019probabilistic,andreas2016neural}. While these approaches provide reliable interpretations by virtue of model design, they are typically restricted to a certain set of modalities or tasks. On the other hand, we propose a more general approach that is able to generate post-hoc explanations for arbitrary black-box multimodal models, and does not assume anything about the modality or classification task itself.

In the latter section on post-hoc explainability of black-box multimodal models, related work has similarly gravitated towards aiming to understand either modality importance~\cite{goyal2016towards,chandrasekaran2018explanations,kanehira2019multimodal} or cross-modal interactions in pretrained language and vision transformer models~\cite{frank2021vision,cao2020behind,parcalabescu2021seeing,li2020does}. Perhaps most related to our work is~\citet{wang2021m2lens} proposing M2Lens, an interactive visual analytics system to visualize and explain black-box multimodal models for sentiment analysis through both unimodal and multimodal contributions. Our approach further disentangles the two types of contributions, which allows us to generate visualizations on each and gain insight into which input features are involved in multimodal interactions. Our approach is also not restricted to sentiment analysis.

\subsection{Representation Disentanglement}

Related to our work is the idea of learning disentangled data representations - mutually independent latent variables that each explain a particular variation of the data~\cite{Bengio:2013:RLR:2498740.2498889,locatello2018challenging}. Disentangled representation learning has been shown to improve both generative and discriminative performance in multimodal tasks~\cite{tsai2018learning}. If the factors of variation are known, many methods learn latent attributes that individually control each variation of data by supervised training~\cite{karaletsos2015bayesian,reed2014learning,cheung2014discovering}. If the factors are partially known or unknown, deep generative models can be used to impose an isotropic Gaussian prior on the latent variables~\cite{vae2013,rubenstein2018latent,Higgins2016VAELB}, maximize the mutual information between a subset of latent variables and the data~\cite{chen2016infogan}, or to encourage the distribution of representations to be factorial and hence independent~\cite{pmlr-v80-kim18b}. Particularly related to our work is empirical multimodally-additive function projection (EMAP)~\cite{hessel2020emap}, an approach for disentangling the effects of unimodal (additive) contributions from cross-modal interactions in multimodal tasks.

\subsection{Dataset and Model Biases}

One core motivation for interpretable ML is to enable a better understanding of the model's decision-making process so as to check whether model behavior is as intended. Using these tools, researchers have uncovered several biases existing in machine learning models and datasets. These biases include undesirable associations captured either in the data or the model, which do not reflect decision-making as one would expect. For example, a line of work in visualizing and understanding multimodal models has uncovered unimodal biases in the language modality of VQA tasks~\cite{jabri2016revisiting,agrawal2016analyzing,anand2018blindfold,cadene2019rubi}, which then inspired follow-up datasets to elevate the importance of visual understanding through VQA 2.0~\cite{goyal2017making}. Similar visualizations also led to improved performance on image captioning tasks by relying less on gender biases and spurious correlations~\cite{hendricks2018women}. Our approach towards better visualizing and understanding multimodal models is also inspired by these insights, and we believe that our fine-grained and general approach will motivate future work towards removing biases from a wider range of datasets and models beyond the prototypical language and vision tasks.

\section{Method: \names}

Our approach, \names\ (short for \namel), is primarily based on disentangling a multimodal model into unimodal contributions (\textbf{UC}) and multimodal interactions (\textbf{MI}), before performing fine-grained visualizations on each disentangled factor. In this section, we introduce precise definitions of unimodal contributions and multimodal interactions, before explaining how disentanglement and interpretations are performed.

\subsection{Unimodal Contributions and Multimodal Interactions}

Unimodal contributions $(\textsc{UC})$ represent information gained by only looking at one of the modalities without interacting with any other modalities, while multimodal interactions $(\textsc{MI})$ are information gained from cross-referencing inputs from multiple modalities~\cite{hessel2020emap}. Multimodal models make decisions using a combination of information from both unimodal contributions and multimodal interactions. For example, in Figure~\ref{intro}, the model assigns a high likelihood to ``glass'' because (1) just by looking at the image, there are many glass objects (unimodal contributions) and (2) by cross-referencing with text, the model focuses on the glass table and assigns a high likelihood to ``glass'' (multimodal interaction).

\begin{figure*}
\vspace{0mm}
\includegraphics[width=1.0\textwidth]{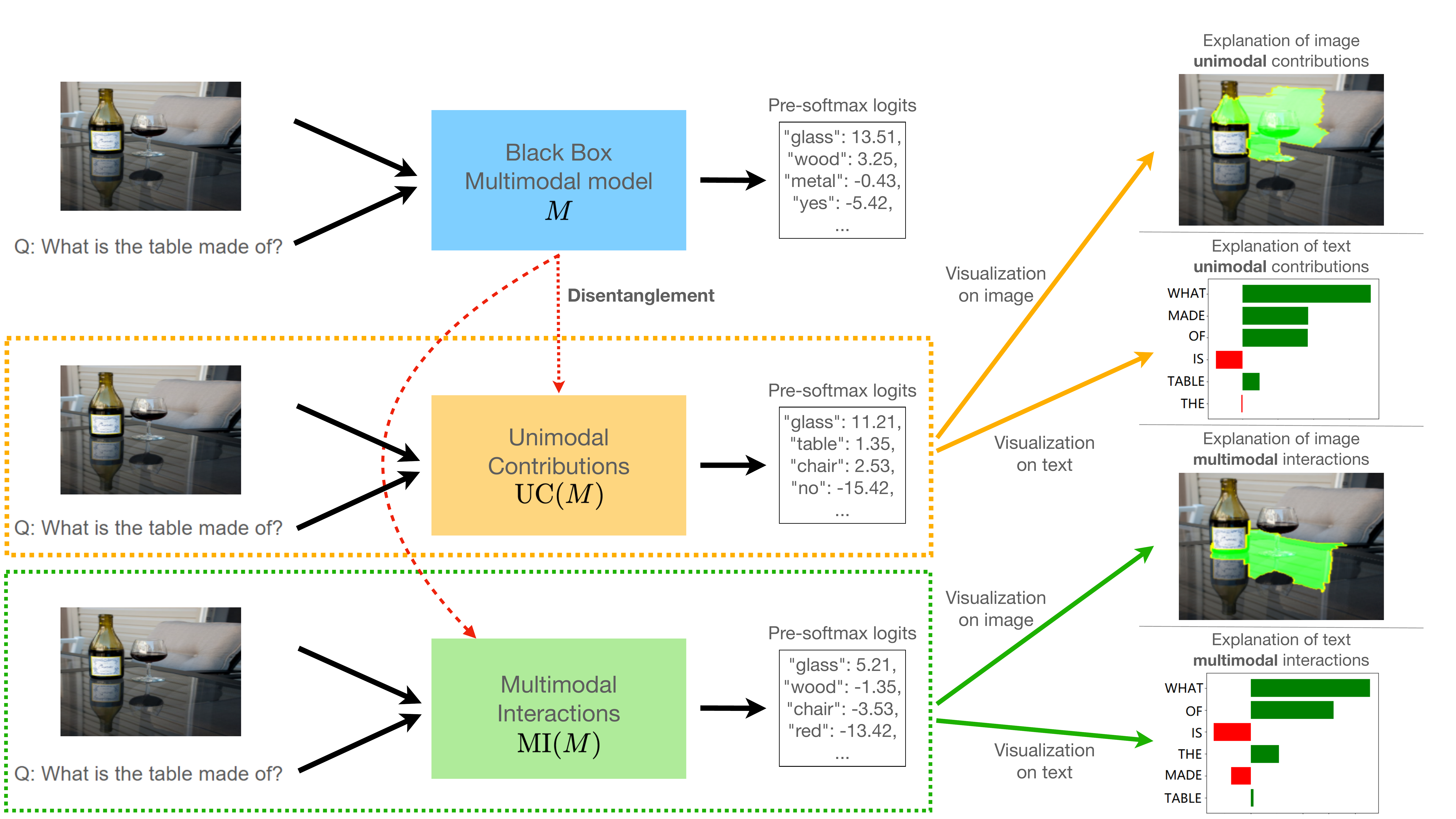}
\caption{High level illustration of \names: we disentangle the model $M$ into two: unimodal contributions (UC) and multimodal interactions (MI), before running visualizations on each sub-model (e.g., using LIME~\cite{lime}) in order to generate fine-grained human-interpretable visualizations of each.}
\label{fig:nonexistent}
\vspace{-3mm}
\end{figure*}

Therefore, to performed fine-grained interpretation in a multimodal model $M$, we first propose a new method to disentangle the model into two submodels:
\begin{equation}
    M = \textsc{UC}(M) + \textsc{MI}(M),
\end{equation}
where $\textsc{UC}(M)$ represents the unimodal contributions within $M$ and $\textsc{MI}(M)$ represents the multimodal interactions within $M$. We can then run visualizations on each sub-model in order to generate human-interpretable visualizations of unimodal contributions and multimodal interactions (see Figure~\ref{fig:nonexistent} for an overview of \names). To generate visual explanations, we choose LIME~\cite{lime}, a widely used interpretation method for black-box models.

\subsection{Model Disentanglement}

\label{sec:proof}

Let $M$ be the multimodal model that we wish to disentangle into unimodal contributions and multimodal interactions. For simplicity, suppose $M$ takes in two modalities as input and produces pre-softmax logits on $C$ classes as output. Therefore, we can view $M$ as a function that maps two inputs $x_1,x_2$ from two modalities to a output logit vector $V$, i.e., $V = M(x_1,x_2)$. Our goal will be to disentangle the function $M$ into a sum of two functions, one representing unimodal contributions and one representing multimodal interactions.

Formally, we would like to write $M$ as $M(x_1,x_2)=g_1(x_1)+g_2(x_2)+g_{12}(x_1,x_2)$, where $g_1$ and $g_2$ are unimodal contributions from the two input modalities, respectively, and $g_{12}$ represents multimodal interactions. By definition of multimodal interactions, we require that $\mathbb{E}_{x_1}g_{12}(x_1,x_2)=0$ for all $x_2$ and $\mathbb{E}_{x_2}g_{12}(x_1,x_2)=0$ for all $x_1$ so that $g_{12}$ contains no unimodal contribution. We will show that under this definition, for each $M$ there will be a unique $g_{12}$ that satisfies these rules. 

We will compute $g_1(x_1)+g_2(x_2)$ using a similar method to EMAP~\cite{hessel2020emap}. We define $\textsc{UC}(M)$ as
\begin{equation}
    \textsc{UC}(M(x_1,x_2)) = \mathbb{E}_{x_1}(M(x_1,x_2)) + \mathbb{E}_{x_2}(M(x_1,x_2)) - \mathbb{E}_{x_1,x_2}(M(x_1,x_2)).
\end{equation}
\textbf{Theorem 1} below (equations 3-5, proof in Appendix) states that $\textsc{UC}(M)$ indeed represents $g_1+g_2$.
\begin{eqnarray}
    && \textsc{UC}(M(x_1,x_2)) \\
    &=& \mathbb{E}_{x_1}(M(x_1,x_2))+ \mathbb{E}_{x_2}(M(x_1,x_2)) - \mathbb{E}_{x_1,x_2}(M(x_1,x_2)) \\
    &=& g_1(x_1)+g_2(x_2).
\end{eqnarray}
Thus, we can compute $g_{12}(x_1,x_2)$ by subtracting $\textsc{UC}(M(x_1,x_2))$ from $M(x_1,x_2)$, which we name $\textsc{MI}(M)$. Formally, 
\begin{eqnarray}
    && \textsc{MI}(M(x_1,x_2)) \\
    &=& M(x_1,x_2) - \textsc{UC}(M(x_1,x_2)) \\
    &=& g_{12}(x_1,x_2).
\end{eqnarray}
This also shows that $g_{12}$ can be uniquely determined.

\begin{figure}
    \centering
    \includegraphics[width=0.45\textwidth]{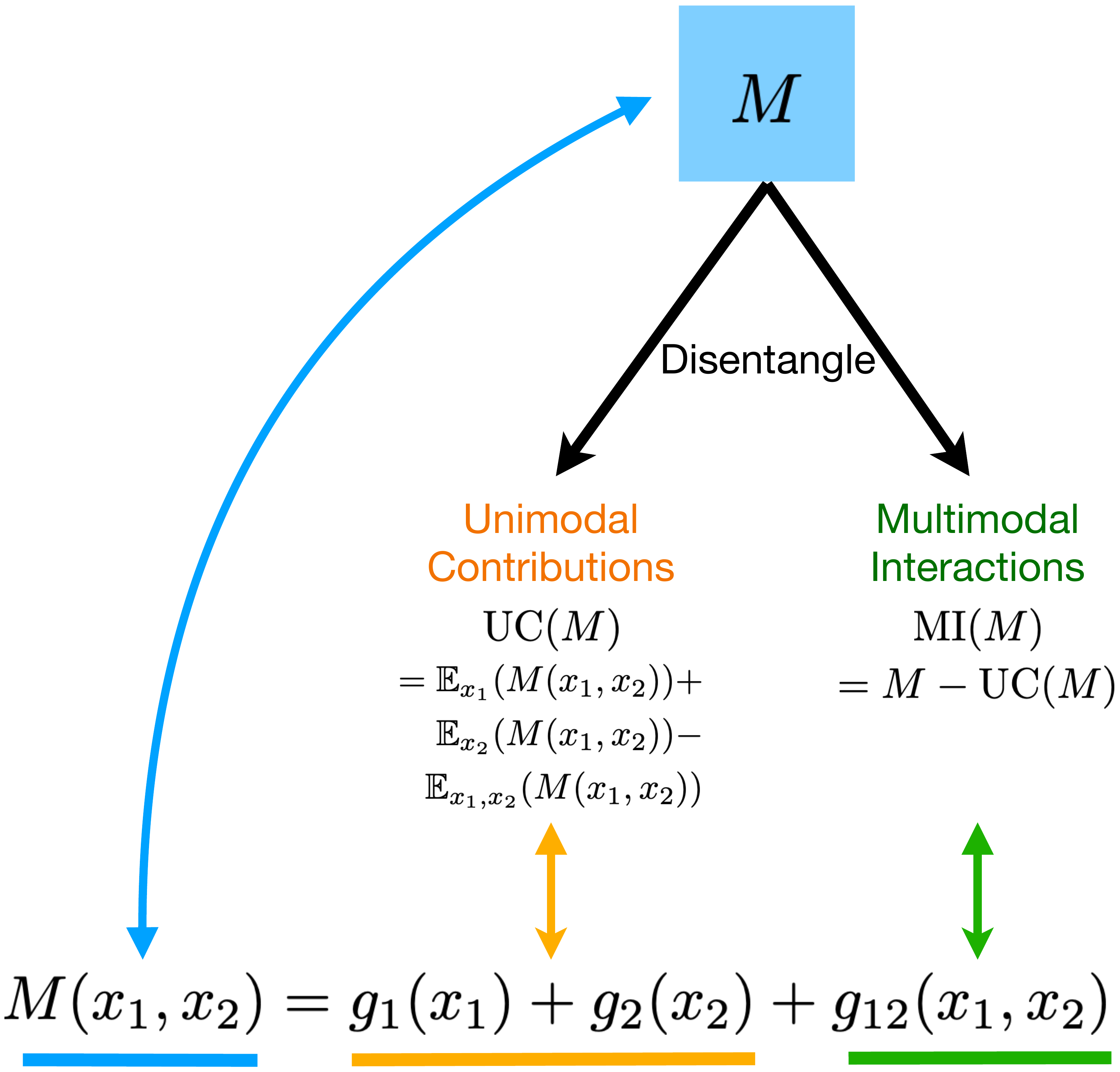}
    \caption{An illustration of the disentangling process of \names. We disentangle a model into two: $\textsc{UC}(M) = g_1 + g_2$ and $\textsc{MI}(M) = g_{12}$, corresponding to unimodal contributions and multimodal interactions respectively.}
    \label{fig:illusdisent}
\end{figure}

In practice, to compute $\textsc{UC}(M(x_1,x_2))$ and $\textsc{MI}(M(x_1,x_2))$, we use a sampling method similar to~\cite{hessel2020emap}, where we sample $N$ datapoints $x^{(i)}=(x^{(i)}_1,x^{(i)}_2)$ including the point we want to explain $x=(x_1,x_2)$ as one of them, and computing each expectation in $\textsc{UC}(M(x_1,x_2))$ by approximating
\begin{align}
    \mathbb{E}_{x_1}(M(x_1,x_2)) &= \sum_{i \in [N]} M(x^{(i)}_1,x_2), \\
    \mathbb{E}_{x_2}(M(x_1,x_2)) &= \sum_{i \in [N]} M(x_1,x^{(i)}_2), \\
    \mathbb{E}_{x_1,x_2}(M(x_1,x_2)) &= \sum_{i \in [N]}\sum_{j \in [N]} M(x^{(i)}_1,x^{(j)}_2).
\end{align}
Figure~\ref{fig:illusdisent} illustrates this disentanglement process.

However, to compute $\textsc{UC}(M(x_1,x_2))$ and $\textsc{MI}(M(x_1,x_2))$, we will need to run forward passes through the model a total of $N^2$ times. In section~\ref{sec:fast} we will show an algorithm that computes this more efficiently by amortizing across multiple datapoints.

\subsection{Interpreting Disentanglement}

Now that we have disentangled the model into two, we will generate human-interpretable explanations on each modality using LIME~\cite{lime}. LIME works by subdividing the input into distinct features, and then randomly perturbing the features $S$ times to see how the perturbations on the features affect the model output logits of a specific class $c$. LIME then fits a linear model mapping the perturbations on each feature to the logits of $c$. The linear model weights on each feature gives the explanation of that feature: if the weight is positive, it means that this feature supports the decision of class $c$; if the weight is negative, it means that this feature is against the decision of class $c$; the larger the weight's absolute value, the stronger the contribution is. Visually, the weights can also be used to generate a human-interpretable visualization: for images, each feature is typically a part of the image, so the parts with the highest absolute weights can be highlighted in green for positive and red for negative contributions. For text, each feature is typically a word, so the explanation can be summarized as a histogram of weights of each word (see Figure~\ref{fig:nonexistent} for an example).

When running LIME on multimodal inputs, we run LIME on one modality at a time, treating the inputs to all other modalities as constant and only perturbing the inputs to that one modality. We denote the generated explanation on model $M$, datapoint $(x_1,x_2)$, and modality $i$ as $\textsc{LIME}_i(M(x_1,x_2))$. After disentanglement into unimodal contributions $\textsc{UC}(M(x_1,x_2))$ and multimodal interactions $\textsc{MI}(M(x_1,x_2))$, our approaches enables the generation of four fine-grained explanations:
\begin{itemize}
    \item $\textsc{UC}_1 = \textsc{LIME}_1(\textsc{UC}(M(x_1,x_2)))$, the explanation of modality 1's unimodal contributions.
    \item $\textsc{UC}_2 = \textsc{LIME}_2(\textsc{UC}(M(x_1,x_2)))$, the explanation of modality 2's unimodal contributions.
    \item $\textsc{MI}_1 = \textsc{LIME}_1(\textsc{MI}(M(x_1,x_2)))$, the explanation of modality 1's contribution to multimodal interactions.
    \item $\textsc{MI}_2 = \textsc{LIME}_2(\textsc{MI}(M(x_1,x_2)))$, the explanation of modality 2's contribution to multimodal interactions.
\end{itemize}

\subsection{Improving Efficiency}
\label{sec:fast}

Since running LIME on a black box model usually requires running the model many times (equal to the LIME sample size $S$), it can be costly to treat $\textsc{UC}(M)$ or $\textsc{MI}(M)$ as black-box models and run LIME on then directly - running $\textsc{UC}(M)$ involves computing $\mathbb{E}_{x_1,x_2}(M(x_1,x_2))$ which requires running $N^2$ forward passes where $N$ is the number of samples used for EMAP, so the total procedure of running \names\ on one datapoint can take $O(SN^2)$ runs of $M$.

In order to make the process faster, we use the following algorithmic trick: we fix $N$ datapoints from the dataset, and then run $M$ on all $N^2$ combinations of the two modalities amongst the $N$ points, and store the resulting logits in a $N\times N\times C$ array $L$ (where $C$ is the number of classes in this task).  When we want to run \names\ on any one of those $N$ points (let's say the $i$th point), for each perturbed LIME sample (WLOG let's say we're running LIME on modality 1, so modality 1 is perturbed in the LIME sample), we make a deep copy of $L$ called $L'$, re-run $M$ on the combination of the perturbed modality 1 input and all $N$ modality 2 inputs, replace the values in the ith row of $L'$ with the results, and compute $\textsc{UC}(M)$ on this LIME sample with the updated table $L'$. Using this trick, after amortizing the one-time initial $O(N^2)$ runs of $M$, each followup \names\ run on any of the $N$ points only takes $O(SN)$ runs of $M$. See details in Algorithm 1 in the Appendix.

\section{Experiments}

In this section, we will perform a set of experiments to fully evaluate the reliability and usefulness of \names\ in interpreting multimodal models. We will be using 3 datasets: a synthetic dataset, CLEVR~\cite{johnson2017clevr}, and VQA 2.0~\cite{balanced_vqa_v2}, and with one corresponding state-of-the-art model for each: MLP, MDETR~\cite{kamath2021mdetr} and LXMERT~\cite{lxmert}. When dealing with datasets involving image and text modalities, we will refer to the two modalities as $(V,T)$ respectively (e.g., $\textsc{UC}_V$ would refer to the \names\ explanation on image unimodal contribution). Our experiments are designed to illustrate the following takeaway messages of using \names\ to analyze multimodal models:
\begin{enumerate}
    \item Our method can reliably disentangle the model and generate accurate explanations for both UC and MI, correlating highly with their respective ground truths (section~\ref{rq1}).
    \item In more difficult tasks such as CLEVR and VQA, and with more complex models, \names\ can still disentangle the model reliably. We show that changing the text input affects $\textsc{UC}_V$ (explanation on image unimodal contribution) little but affects $\textsc{MI}_V$ (explanation on multimodal interactions from the image side) significantly (section~\ref{rq1}).
    \item \names\ gives additional insight into understanding multimodal model behavior by answering whether the model relies mostly on UC, MI, or both in making the prediction (section~\ref{rq2}).
    \item \names\ also enables human users to debug and improve models by identifying which input features are used in MI and revealing undesirable behavior in models (section~\ref{rq3}).
\end{enumerate}
Following these results, we will discuss limitations and future works (section~\ref{limit}). 

\subsection{Setup}

\subsubsection{Datasets}

We will use three datasets: a synthetic dataset to enable controlled variations between unimodal and multimodal interactions, as well as two large-scale multimodal datasets: CLEVR, and VQA 2.0.

The \textbf{synthetic dataset $D$} is designed to model a task that requires both unimodal (additive) contributions and multimodal interactions to solve correctly. According to prior work~\cite{hessel2020emap}, the dot product of two modalities requires non-additive cross-modal interaction, while the sum of two vectors is additive. Therefore, we design a synthetic dataset $D$ by randomly generating two $10$-dimensional vectors following $N(0,1)$ independently for each element, and then computing the sum of all elements in both vectors plus the dot product of the two vectors. If the result's absolute value is below $0.01$, we discard this point; otherwise, we assign a $0/1$ label based on the sign of the result. We generate $100,000$ points to form $D$ and divide it into train/valid/test splits by $8/1/1$ ratio.

\textbf{CLEVR}~\cite{johnson2017clevr} is a diagnostic dataset designed for language and visual reasoning. The dataset consists of synthesized images of 3D shapes of various colors, sizes, and materials on a gray background, For each image, there are several questions about the shapes' attributes, positions, and numbers. This dataset has been widely used for diagnostic purposes to find model weaknesses.

\textbf{VQA 2.0}~\cite{balanced_vqa_v2} is a dataset containing various questions on real-world images. It is designed to force multimodal interactions, especially incorporating the visual aspect, by sometimes having the same question with two different answers on two different images. This dataset is interesting because models have been shown to occasionally ``guess'' correct answers purely from unimodal contributions or with the wrong visual grounding~\cite{cadene2019rubi,anand2018blindfold}. \names\ will enable us to study how often models rely on undesirable unimodal biases and further understand the model's decision-making process.

\subsubsection{Models}

For synthetic dataset $D$, we train a \textbf{4-layer MLP} (with input size $20$ and hidden layer sizes $100,200,10,2$ respectively) on $D$ that reaches $97.3\%$ accuracy on the test split.

For CLEVR dataset, we will be using a pretrained \textbf{MDETR}~\cite{kamath2021mdetr} that achieves $99.7\%$ test accuracy. 

For VQA 2.0, we will be using pretrained \textbf{LXMERT}~\cite{lxmert}, one of the best models on the dataset, with a $72.5\%$ test accuracy.

\subsection{Research Questions and Results}

\begin{table*}[]
\vspace{2mm}
\begin{tabular}{l|ccc | ccc}
    \Xhline{3\arrayrulewidth}
    Ground Truth Explanations & $\names\ \textrm{UC}_1$ & $\names\ \textrm{MI}_1$ & $LIME_1(M)$ & $\names\ \textrm{UC}_2$ & $\names\ \textrm{MI}_2$ & $LIME_2(M)$  \\
    \hline
    Modality 1 Unimodal Bias ($d_1$) & $\mathbf{0.982}$ & $0.027$ & $0.700$ & $-0.004$ & $0.002$ & $0.003$ \\ 
    Modality 2 Unimodal Bias ($d_2$) & $-0.005$ & $-0.005$  & $-0.006$ & $\mathbf{0.979}$ & $-0.003$ & $0.675$ \\
    Multimodal Interaction ($d_1*d_2$) & $-0.001$  & $\mathbf{0.960}$ & $0.627$ & $-0.003$ & $\mathbf{0.947}$ & $0.654$ \\
    \Xhline{3\arrayrulewidth}
\end{tabular}
\vspace{1mm}
\caption{Pearson correlation between the LIME explanation vectors and ground-truth explanations for $d_1$, $d_2$, and element-wise product $d_1*d_2$. With disentanglement in \names, the unimodal contribution explanations completely correlates with their respective unimodal ground truth, and the multimodal interaction explanations completely correlates with the ground truth multimodal interactions (i.e., element-wise product of the two inputs). On the other hand, running LIME without disentangling gives an explanation that confuses both unimodal contributions and multimodal interactions.}
\label{tab:synth}
\vspace{-3mm}
\end{table*}

\begin{table}[]
    \centering
    \begin{tabular}{l|cc}
        \Xhline{3\arrayrulewidth}
        Dataset (Model) & $\textrm{UC}_V$ & $\textrm{MI}_V$ \\
        \hline
        CLEVR (MDETR)  & $0.005$ & $0.295$ \\
        VQA (LXMERT) &  $0.001$ & $0.808$ \\
        \Xhline{3\arrayrulewidth}
    \end{tabular}
    \vspace{1mm}
    \caption{Average cosine distance between \names\ image explanations before/after text swap. The result shows that unimodal image contributions explanations ($\textrm{UC}_V$) are almost not affected by changes in the text modality, while multimodal interaction explanation from the image side ($\textrm{MI}_V$) is affected by changes in the other modality significantly, so the disentangling effect of \names\ works as intended.}
    \label{tab:relia}
    \vspace{-5mm}
\end{table}

\subsubsection{\textbf{RQ1:} Can \names\ reliably disentangle a model into unimodal contributions and multimodal interactions and generate accurate explanations for both UC and MI in practice?}
\label{rq1}

\ 

In section~\ref{sec:proof}, we have theoretically shown that \names\ can disentangle a model into unimodal contributions and multimodal interactions. To show that this also holds in practice (when expectation computations are replaced by sampling), we will run \names\ on our trained model $M$ using $1,000$ randomly selected datapoints in the test split of our synthetic dataset $D$, on label $1$ (i.e., that the sum of all elements of both vectors plus the dot-product of the two vectors are positive).

For each point $(d_1,d_2)$ in $D$, since we are classifying whether the sum of all elements in $d_1$ and $d_2$ as well as the dot product of $d_1$ and $d_2$, the ground truth UC explanation on each modality will be $d_1$ and $d_2$ respectively, and the ground truth MI explanation will be element-wise product $d_1*d_2$. Therefore, for each generated explanation on input data $(d_1,d_2)$, we will compute the Pearson Correlation between the explanation weights of the $10$ features with the values of the $10$ features of $d_1$, the values of the $10$ features of $d_2$, and the $10$ features in the element-wise product of $d_1$ and $d_2$. In addition to \names, we also run LIME under the same settings as an ablation and compute average correlations.

The results are shown in Table~\ref{tab:synth}. We found that within each datapoint ($d_1$,$d_2$), there is a strong correlation between each \names-generated unimodal explanation ($\textsc{UC}_1, \textsc{UC}_2$) and the corresponding ground truth UC explanation, but there is neither correlation between $\textsc{UC}_1$/$\textsc{UC}_2$ and ground truth UC explanation of a different modality, nor correlation between $\textsc{UC}_1$/$\textsc{UC}_2$ and ground truth multimodal interaction explanations. This shows that \names-generated UC explanations indeed capture unimodal contributions only. Moreover, we found that both \names-generated multimodal interaction explanations ($\textsc{MI}_1, \textsc{MI}_2$) indeed correlate with the ground truth MI explanation, but not with either ground truth UC explanation. This shows that \names-generated multimodal interaction explanation indeed captures explanations on just the multimodal interactions (i.e., the dot-product), and not any of the unimodal contributions. Meanwhile, running the original LIME on either modality just gives an explanation that weakly correlates with ground truth unimodal contributions and multimodal interactions, so the original LIME without disentangling is unable to give an accurate explanation of either unimodal contributions or multimodal interactions.

In addition to using a synthetic dataset, we show that \names\ can also disentangle more complex models on multimodal tasks, such as MDETR on CLEVR and LXMERT on VQA (the latter model is far from perfect in performance). As a measure of disentanglement, we check how \names-generated explanations would be different given the same image but different questions. From each dataset, we randomly select $100$ points and generate their \names\ explanations on the correct label. Then, for each point, we swap out the question with another different question on the same image and generate their \names\ explanations on the same label (i.e., correct label before the swap). We compute cosine distance between the explanation weights from $\textsc{UC}_V$ before/after the swap, as well as cosine distance between the weights from $\textsc{MI}_V$ before/after the swap, and report average cosine distances on each dataset in Table~\ref{tab:relia}. We can see that swapping text has almost no effect on $\textsc{UC}_V$ but affects $\textsc{MI}_V$ significantly. Therefore, \names\ is able to correctly disentangle a model into unimodal contributions and multimodal interaction for more complex models and tasks.

\subsubsection{\textbf{RQ2:} Can \names\ help researchers gain additional insight in whether unimodal contributions or multimodal interactions are the dominant factors behind a model's prediction?}
\label{rq2}

\ 

Disentangling the model into UC and MI and generating visualizations for each should provide additional insights into whether UC or MI is the main factor in the model's prediction. In the following experiments, we show that \names\ can uncover which factor is dominant in a model's prediction process both across all points in the dataset (``global'') and on each individual datapoint (``local'').

\textbf{Global interpretation:} CLEVR dataset is designed to force multimodal interactions, and MDETR has a $99.7\%$ accuracy on CLEVR, so we expect that MDETR will be heavily reliant on multimodal interactions. To verify this, we run \names\ on MDETR for $100$ randomly sampled datapoints from the validation split of CLEVR, and compute the average absolute weight of the top-5 features in \names\  explanations. As shown in Table~\ref{tab:weight}, the $\textsc{MI}_V$ and $MV_T$ weights are indeed significantly larger than $\textsc{UC}_V$ and $\textsc{UC}_T$ weights. Note that unimodal text does still give some useful information in CLEVR, such as the answer type (yes/no, attribute, or number), so that explains why $\textsc{UC}_T$ still has a weight of about $60\%$ that of $\textsc{MI}_T$. The average weight for $\textsc{MI}_V$, however, is over $4$ times higher than $\textsc{UC}_V$. Therefore, using \names, we confirmed that MDETR indeed relies mostly on multimodal interactions to solve the task.

\begin{table}[]
    \centering
    \begin{tabular}{l|cc}
        \Xhline{3\arrayrulewidth}
        & Text & Image\\
        \hline
        Unimodal contributions & ($\textrm{UC}_T$): $2.31$ & ($\textrm{UC}_V$): $0.38$ \\
        Multimodal interactions & $\mathbf{(MI_T): 3.81}$  & $\mathbf{(MI_V): 1.63}$ \\
        \Xhline{3\arrayrulewidth}
    \end{tabular}
    \vspace{1mm}
    \caption{Average absolute weight of top-5 features in \names\ explanations using MDETR model on CLEVR dataset. The average MI explanation weights are much larger than the average UC explanation weights. This shows that multimodal interaction is the dominant factor that MDETR relies on when making predictions for the overwhelming majority of datapoints.}
    \label{tab:weight}
    \vspace{-3mm}
\end{table}

\textbf{Local interpretation:} In most datasets and models, models will not be near-perfect, and they will have different dominating factors from datapoint to datapoint. In this case, a global analysis will not suffice, and it will be necessary to look into which factor contributes more to the model's prediction on individual datapoints. We perform the following experiment to show that \names\ can help users determine whether a model makes a prediction on a datapoint where (1) unimodal text is dominant, (2) unimodal image is dominant, (3) multimodal interactions are dominant, and (4) both UC and MI have significant contributions to the answer. We will use LXMERT on VQA since LXMERT is not close to perfect and often relies on different factors when predicting different datapoints. 

We gave five human annotators (who have some background knowledge in machine learning but do not have any knowledge about \names) the same set of $52$ datapoints from VQA, as well as the prediction from LXMERT. For each datapoint, each human annotator is first given the LIME explanations without disentanglement as a baseline, and they are asked to categorize this point into one of the four categories above, while also rating how confident they are on their decision on a scale from one (least confident) to five (most confident). The human annotators are then presented with \names\ explanations, and again they are asked to categorize each point as well as rate their confidence.

The results are shown in Table~\ref{tab:newanno}. We can see that human annotators have significantly higher average confidence scores when presented with \names\ explanations as compared to the baseline. Moreover, \names\ result shows significantly higher Krippendorff's alpha score~\cite{krippendorff2011computing}, which measures inter-annotator agreements, so annotators also tend to agree a lot more on their categorizations. Therefore, \names\ is able to help researchers more confidently determine whether UC or MI (or both) is the dominant factor behind the model's prediction, and thus help researchers gain additional insight into model behavior.

\begin{table}[]
    \centering
    \vspace{2mm}
    \begin{tabular}{l|cc}
    \Xhline{3\arrayrulewidth}
& LIME & DIME \\ \hline
Average confidence score & $2.23$ & $\mathbf{3.77}$ \\
Annotator agreement $(\textrm{Krippendorff's } \alpha)$~\cite{krippendorff2011computing} & $0.18$ & $\mathbf{0.57}$ \\
\Xhline{3\arrayrulewidth}
\end{tabular}
\vspace{1mm}
\caption{Results of the human annotation experiments on categorizing whether LXMERT predicts a point using UC, MI, or both as the dominant deciding factor. We can see that on average, human annotators gave much higher confidence score to the \names\ explanation compared to LIME without disentanglement, and human annotators also tend to agree more on decisions based on \names.}
    \label{tab:newanno}
    \vspace{-5mm}
\end{table}

\subsubsection{\textbf{RQ3}: Can \names\ help us better assess the qualities of the model and gain insights on how to debug or improve model performance?}
\label{rq3}

\ 

When trying to debug or improve a model on a task involving challenging reasoning, such as VQA, one important question researchers often ask is: do we know if our model actually learns to do the task ``the intended way'' (i.e., go through the same logical reasoning process as a human would to perform the task)? How often does our model perform as intended? Therefore, we conduct the following experiment to show that \names\ may help answer this question.

We use \names\ explanations to categorize the model's behavior on each datapoint into one of the following categories:

When the model answers correctly,
\begin{itemize}

    \item (1) The model fully identifies the necessary parts of the image to answer the question logically through MI.

    \item (2) The model only partially identifies the parts of the image that are necessary to answer the question logically through MI and got it right with help of unimodal contributions.

    \item (3) The model did not correctly identify any of the parts of the image that are necessary to answer the question logically through MI. It got it right purely by unimodal contributions or by chance.

\end{itemize}

And when the model answers incorrectly,
\begin{itemize}
    \item (4) The model fully identifies the necessary parts of the image to answer the question logically through MI, but still gets the answer wrong because the model does not fully understand a concept or because the question is too difficult (even for a human being).
    \item (5) The model only partially identifies the parts of the image that are necessary to answer the question logically through MI, thus missing some of the key parts of the image resulting in an incorrect answer.
    \item (6) The model did not correctly identify any of the parts of the image that are necessary to answer the question logically through MI, and thus the model fails to answer the question correctly.
\end{itemize}

\begin{figure*}
\vspace{2mm}
\includegraphics[width=1.0\textwidth]{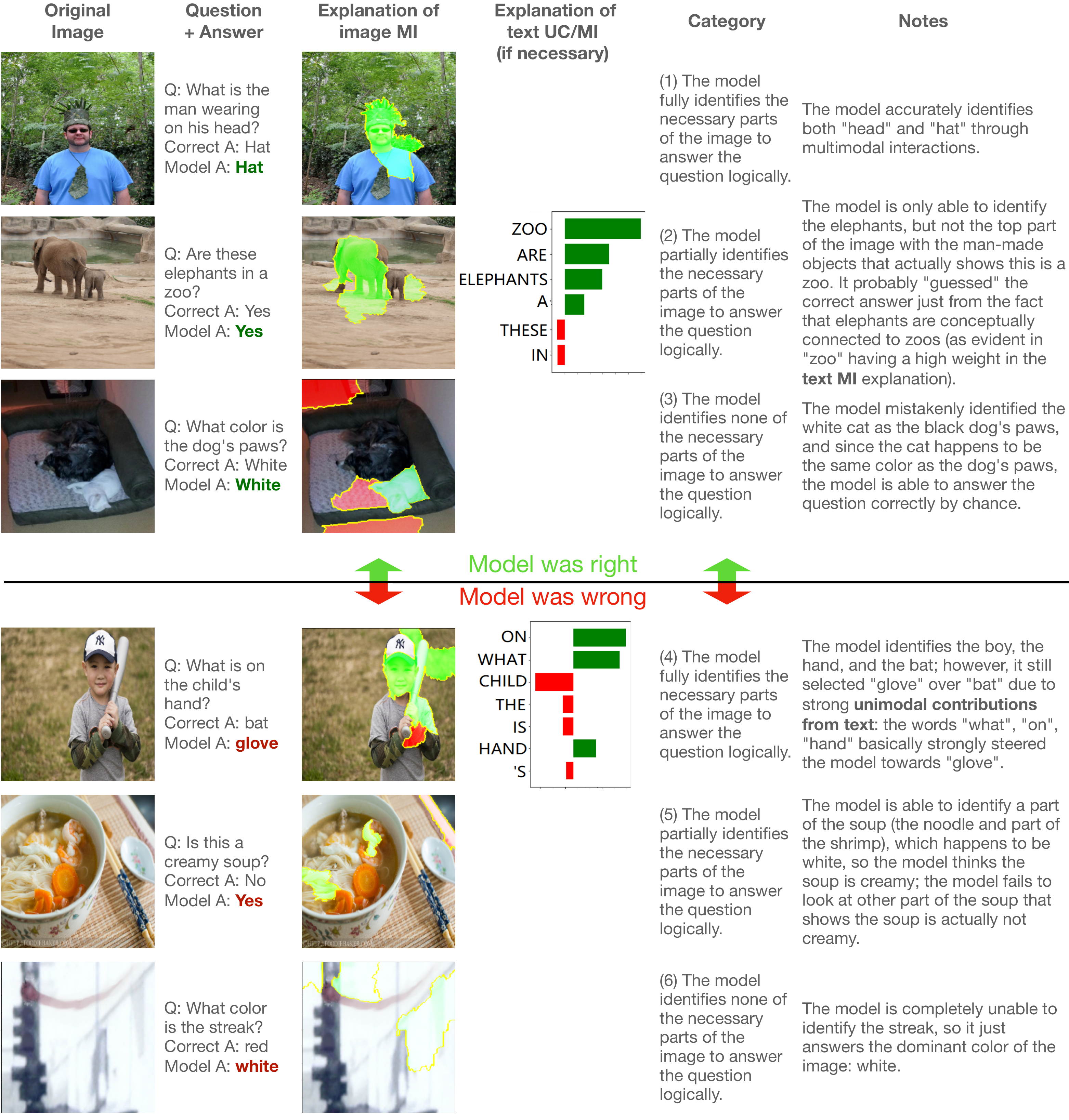}
\caption{Here we present examples of using \names\ to categorize and explain why LXMERT makes certain predictions on datapoints in VQA 2.0. We present one example from each category. In most cases, only looking at the multimodal interaction explanations from the image side ($\textsc{MI}_V$) is sufficient to explain and categorize the model, but in certain cases, additional information from $\textsc{UC}_V$, $\textsc{UC}_T$, or $\textsc{MI}_T$ is needed as well. \names\ enables researchers to gain understanding of the model's decision-making process which presents a step towards debugging and improving these models.}
\label{fig:veryinterestingexamples}
\end{figure*}

In Figure~\ref{fig:veryinterestingexamples}, we show examples of datapoints, model predictions, and explanations that were annotated into each of the above categories. As shown in the examples, in most cases, there will be enough evidence to categorize a datapoint just by looking at the multimodal interaction explanations from the image side ($\textsc{MI}_V$), but sometimes other \names\ explanations (e.g., explanations of text interactions) will be needed to gain additional understanding of the model's decision-making process. 

\begin{table}[]
    \centering
    \vspace{2mm}
    \setlength\tabcolsep{4.0pt}
    \begin{tabular}{l|cccc}
        \Xhline{3\arrayrulewidth}
        & \multirow{2}{*}{Total} & Fully & Partial & Not at all \\ 
        & & (Cat. $1,4$) & (Cat. $2,5$) & (Cat. $3,6$) \\
        \hline
        Total & $118$ & $69$ & $35$ & $34$ \\
        LXMERT Correct & $87$ & $59$ & $26$ & $22$ \\
        LXMERT Incorrect & $31$ & $10$ & $9$ & $12$ \\
        \Xhline{3\arrayrulewidth}
    \end{tabular}
    \vspace{1mm}
    \caption{We asked human annotators to categorize data points from VQA with evidence from \names. While LXMERT is able to fully identify the necessary regions of the image half of the time $(69/118)$, there is still a significant portion of datapoints where LXMERT is unable to fully identify the necessary regions and relies on either guessing or on unimodal contributions.\vspace{-5mm}}
    \label{tab:vqa}
\end{table}

The results of this human study are shown in Table~\ref{tab:vqa}. With \names, we were able to categorize $118$ points with evidence, out of a total of $140$ points $(84\%)$. This shows that \names\ is able to highlight which input features are aligned or recognized by MI. We observe that, even though the models can fully identify the correct parts of the image that are relevant to the questions half of the time $(69/118)$, there is still a significant portion of datapoints where the model correctly aligns text and image but relies on unimodal contributions instead. This highlights several shortcomings of the model's decision-making process despite answering the question correctly. Therefore, the information gained from performing \names\ can help researchers identify weaknesses in their models and debug or improve these models accordingly.

We also observe that the model is more likely to not be able to fully identify the correct regions of the image when the model makes the wrong prediction, which is expected.

\begin{figure}
    \centering
    \vspace{2mm}
    \includegraphics[width=0.44\textwidth]{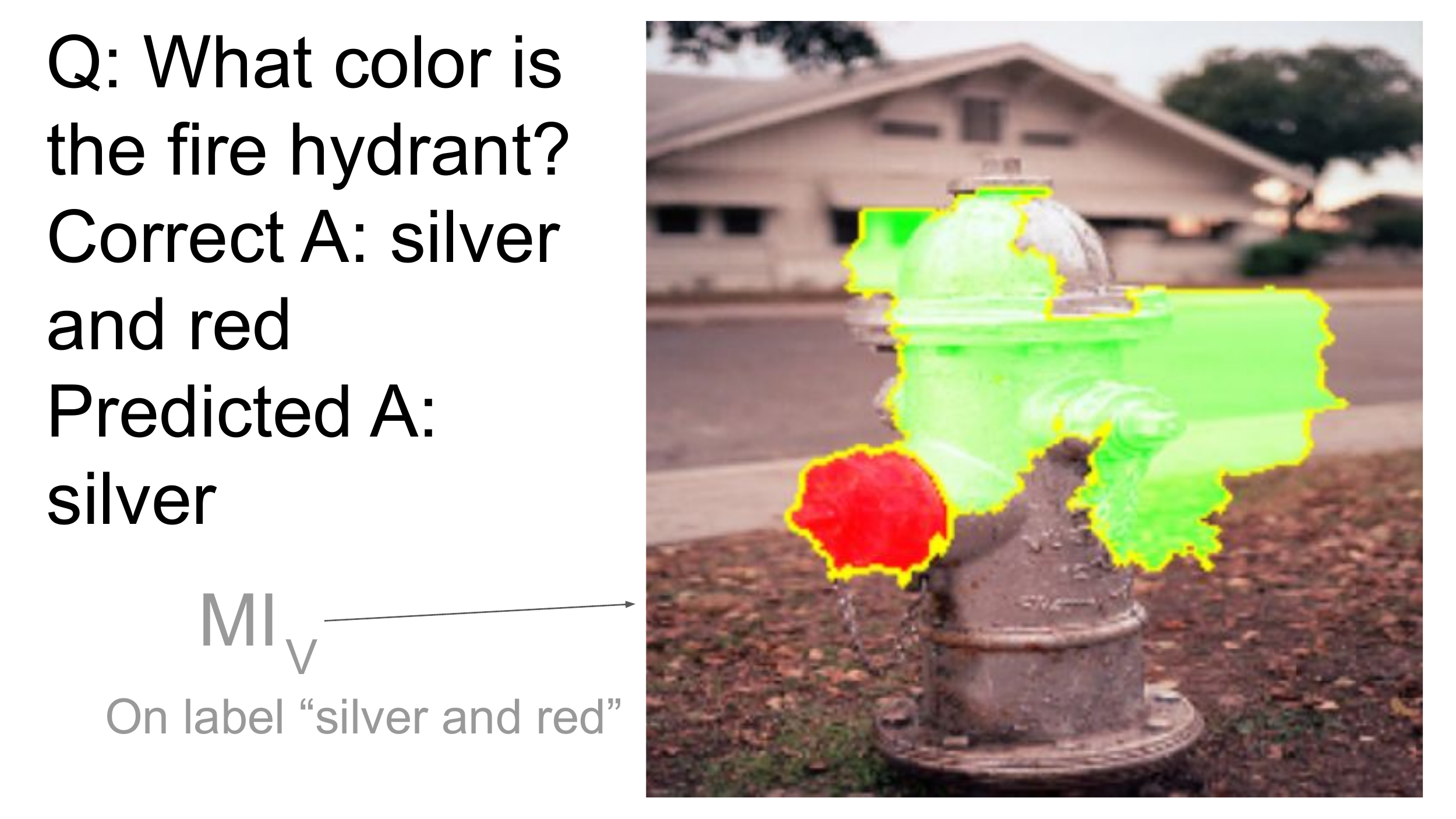}
    \caption{In this example, the model was unable to answer correctly because it did not recognize the red part in the image as part of the hydrant. As shown by the $\textsc{MI}_V$ explanation, the model actually thought that the red part is ``against'' the answer ``silver and red'', which means the model thought the red region isn't a part of the hydrant.}
    \label{fig:hydrant}
\end{figure}

In addition, we also found the following interesting observations when looking at the \names\ explanations of the $118$ points:

\begin{itemize}
    \item  LXMERT often relies too heavily on unimodal text contributions: for example, in a question involving ``car'', unimodal contributions in text will prompt the model to answer ``street'' even if the model is unable to find ``street'' in the image. Sometimes, even when the model is able to interactively identify the correct regions of the image, unimodal text contributions can still dominate over the multimodal interaction (such as the fourth example in Figure~\ref{fig:veryinterestingexamples}, where the model answered ``glove'' due to unimodal text contributions even though the model was able to interactively identify the bat). 
    \item The model sometimes interactively identifies the wrong object that happens to share the same properties in question as the correct object (such as the third example in Figure~\ref{fig:veryinterestingexamples}, where instead of the dog's paws, the model identified the nearby cat which also happens to be white). This coincidence happens more often than we expected, as there are $8$ such cases amongst the $118$ examples $(7\%)$.
    \item When asked about the color of an object that has two colors, LXMERT will only pick out one of the colors. \names\ analysis shows that this is often due to LXMERT only identifying subregions of the object in one color while ignoring other parts of the object that are in a different color. For example, in Figure~\ref{fig:hydrant}, the model thinks that the hydrant is not ``silver and red'' because it did not classify the red tip as part of the hydrant.
\end{itemize}

These additional observations may guide future research in improving LXMERT (and other similar models) or designing inductive biases to avoid these undesirable behaviors.

\subsection{Limitations and Future Directions}
\label{limit}

Despite the ability of \names\ in interpreting and debugging multimodal models, there remain several directions for future work:

\textbf{1. Models with discrete outputs:} Even though \names\ is designed to work for any black-box classification models, it requires the model to produce a continuous logit for each answer choice. \names\ does not work well on the Neural-Symbolic VQA model~\cite{Mao2019NeuroSymbolic} since it only produces one discrete output instead of a continuous logit. Even when we tried to convert its outputs to logits by assigning its answer a logit of 1 and all other answer choices a logit of $-1$, \names\ often fails to produce any meaningful explanation since the perturbations are unable to change the discrete answer of the model, thus having no effect on the assigned logits.

\textbf{2. Number of modalities:} In all experiments, \names\ was applied to tasks with 2 modalities. Disentangling a model across even 3 modalities can be very costly, as we will need to run the model $N^3$ times to compute unimodal contributions. Another challenge lies in interpreting the multimodal interaction, which would consist of bi-modal interactions between each pair of modalities as well as tri-modal interactions across all 3 modalities. Future work should tackle these challenges and try to expand \names\ for high-modality scenarios.

\textbf{3. Diverse modalities:} Even though the disentangling method in \names\ theoretically works on any modality, our experiments have focused on image+text datasets (except the synthetic dataset experiment). This is because LIME-generated visualized explanations are relatively intuitive on image and text; it can be much harder for a human annotator to look at the results of explanations on other modalities (such as time-series of vectors) and try to make sense of them. In the future, we would like to design additional experiments to show that \names\ can also be used to gain additional insight on model behavior in tasks involving modalities other than image and text as well.

\textbf{4. Using these insights to improve models:} Since \names\ is able to reveal several hidden undesirable behaviors in multimodal models, future work should aim to propose targeted solutions to these highlighted biases as a step towards improving multimodal models. For example, according to insights gained on VQA in RQ3, LXMERT can be improved by encouraging less reliance on unimodal text contribution, where insights from~\citet{cadene2019rubi} (which studies this research question for non-pretrained models) could be useful. Furthermore, future work could also design new training objectives which penalize models that associate wrong objects with words in MI, despite getting the correct answer.

\section{Conclusion}

In conclusion, \names\ presents a new way to help users understand multimodal models by disentanglement into unimodal contributions and multimodal interactions before generating visual explanations for each. \names\ can generate accurate disentangled explanations, help researchers and developers gain a deeper understanding of model behavior, and presents a step towards debugging and improving these models. We hope that \names\ inspires the design of multimodal models that are more trustworthy, reliable, and robust for real-world applications.

\section*{Acknowledgements}

This material is based upon work partially supported by the National Science Foundation (Awards \#1722822 and \#1750439) and National Institutes of Health (Awards \#R01MH125740, \#R01MH096951, and \#U01MH116925). 
PPL is partially supported by a Facebook PhD Fellowship and a Carnegie Mellon University's Center for Machine Learning and Health Fellowship. RS is partially supported by NSF IIS1763562 and ONR Grant N000141812861.
Any opinions, findings, conclusions, or recommendations expressed in this material are those of the author(s) and do not necessarily reflect the views of the National Science Foundation, National Institutes of Health, Facebook, Carnegie Mellon University's Center for Machine Learning and Health, or Office of Naval Research, and no official endorsement should be inferred. We are extremely grateful to Gunjan Chhablani, Martin Ma, Chaitanya Ahuja, Volkan Cirik, Peter Wu, Amir Zadeh, Alex Wilf, Victoria Lin, Dong Won Lee, and Torsten W\"{o}rtwein for helpful discussions and feedback on initial versions of this paper. Finally, we would also like to acknowledge NVIDIA's GPU support.

\bibliography{refs}
\bibliographystyle{ACM-Reference-Format}

\clearpage

\onecolumn

\appendix

\section{Proof of Theorem 1}

\textbf{Theorem 1} below states that $\textsc{UC}(M)$ indeed represents $g_1+g_2$.
\begin{eqnarray}
    && \textsc{UC}(M(x_1,x_2)) \\
    &=& \mathbb{E}_{x_1}(M(x_1,x_2))+ \mathbb{E}_{x_2}(M(x_1,x_2)) - \mathbb{E}_{x_1,x_2}(M(x_1,x_2)) \\
    &=& g_1(x_1)+g_2(x_2)
\end{eqnarray}

\textbf{Proof:}
\begin{eqnarray}
&& \textsc{UC}(M(x_1,x_2)) \\
&=& \mathbb{E}_{x_1}(M(x_1,x_2)) + \mathbb{E}_{x_2}(M(x_1,x_2)) - \mathbb{E}_{x_1,x_2}(M(x_1,x_2)) \\
&=& \mathbb{E}_{x_1}(g_1(x_1)+g_2(x_2)+g_{12}(x_1,x_2))+\mathbb{E}_{x_2}(g_1(x_1)+g_2(x_2)+g_{12}(x_1,x_2))-\mathbb{E}_{x_1,x_2}(g_1(x_1)+g_2(x_2)+g_{12}(x_1,x_2)) \\
&=& \mathbb{E}_{x_1}(g_1(x_1))+\mathbb{E}_{x_1}(g_2(x_2))+\mathbb{E}_{x_2}(g_1(x_1))+\mathbb{E}_{x_2}(g_2(x_2))-\mathbb{E}_{x_1,x_2}(g_1(x_1))-\mathbb{E}_{x_1,x_2}(g_2(x_2)) \\
&=& \mathbb{E}_{x_1}(g_1(x_1))+g_2(x_2)+\mathbb{E}_{x_2}(g_2(x_2))+g_1(x_1)-\mathbb{E}_{x_1}(g_1(x_1))-\mathbb{E}_{x_2}(g_2(x_2)) \\
&=& g_1(x_1)+g_2(x_2)
\end{eqnarray}

\section{Algorithm Details}

In Algorithm~\ref{alg:code} we describe our procedure for efficiently running batched \names. The core idea in \names\ is to provide more fine-grained interpretations by disentangling a multimodal model into unimodal contributions (\textbf{UC}) and multimodal interactions (\textbf{MI}). \names\ is able to accurately perform disentanglement and generate reliable explanations for both UC and MI. Using \names, we are able to gain additional insights on model behavior and better determine whether UC, MI, or both are the dominant factor behind the model's predictions on individual datapoints. Furthermore, \names\ presents a step towards debugging and improving these models as it may reveal certain undesirable behaviors of the models.

\begin{algorithm}
    \vspace{2mm}
    \caption{Efficient algorithm for running batched \names.}
    \label{alg:code}
     \begin{algorithmic}
    
      \STATE \textbf{DIME}($M,X,S,d,\textrm{PERTURB}$)
      
      \STATE (Where $M$ is our model that returns pre-softmax logits, $X=(x_1^{(i)},x_2^{(i)},c^{(i)})_{i\in [N]}$ is a set of $N$ points, and $N$ is the number of samples needed for approximating expectations when computing UC, $S$ is the sample size for LIME, $C$ is the number of classes)
      
      \STATE ($\textrm{PERTURB}$ is a function where $\textrm{PERTURB}(x,S)$ returns S random LIME perturbations on x)
      
      \STATE $L = 0_{N\times N\times C}$
        \FOR{$i = 0$ to $N-1$}
            \FOR{$j = 0$ to $N-1$}
            \STATE $L[i][j] = M(x_1^{(i)},x_2^{(j)}) $
            \ENDFOR
        \ENDFOR
        
        \STATE (Run on first modality)
        
        \FOR{$k = 0$ to $N-1$} 
            \STATE $\textrm{pred} = \textrm{argmax}(\textrm{cache}[k][k])$
            \STATE $Z = \textrm{PERTURB}(x_1^{(k)},S)$
            
            \STATE $\textrm{uni} = 0_{S\times C}$
            \STATE $\textrm{multi} = 0_{S\times C}$
            \FOR{$s = 0$ to $S-1$}
            \STATE $L' = \textrm{deepcopy}(L)$
              \FOR{$n = 0$ to $N-1$}
              \STATE $L'[k][n] = M(Z[s],x_2^{(n)})$
              \ENDFOR
              \STATE $\textrm{avg}_1 = \textrm{AVG}(L'[k], \textrm{dim}=0)$
              \STATE $\textrm{avg}_2 = \textrm{AVG}(L'[:,k], \textrm{dim}=0) $
              \STATE $\textrm{avg}_{12} = \textrm{AVG}(L', \textrm{dim}=\{0,1\})$
              \STATE $\textrm{uni}[s] = \textrm{avg}_1 + \textrm{avg}_2 - \textrm{avg}_{12}$
              \STATE $\textrm{multi}[s] = L'[k][k] - \textrm{uni}[s]$
            \ENDFOR
            \STATE $\textsc{UC}_1 = \textsc{LIME}_1(Z, \textrm{uni})$
            \STATE $\textsc{MI}_1 = \textsc{LIME}_1(Z, \textrm{multi})$
        \ENDFOR
        
        \STATE (Run on second modality)
        
        \FOR{$k = 0$ to $N-1$} 
            \STATE $\textrm{pred} = \textrm{argmax}(\textrm{cache}[k][k])$
            \STATE $Z = \textrm{PERTURB}(x_2^{(k)},S)$
            
            \STATE $\textrm{uni}=0_{S\times C}$
            \STATE $\textrm{multi}=0_{S\times C}$
            \FOR{$s = 0$ to $S-1$}
            \STATE $L' = \textrm{deepcopy}(\textrm{cache})$
              \FOR{$n = 0$ to $N-1$}
              \STATE $L'[n][k] = M(x_1^{(n)},Z[s])$
              \ENDFOR
              \STATE $\textrm{avg}_1 = \textrm{AVG}(L'[k], \textrm{dim}=0)$
              \STATE $\textrm{avg}_2 = \textrm{AVG}(L'[:,k], \textrm{dim}=0) $
              \STATE $\textrm{avg}_{12} = \textrm{AVG}(L', \textrm{dim}=\{0,1\})$
              \STATE $\textrm{uni}[s] = \textrm{avg}_1 + \textrm{avg}_2 - \textrm{avg}_{12}$
              \STATE $\textrm{multi}[s] = L'[k][k] - \textrm{uni}[s]$
            \ENDFOR
            \STATE $\textsc{UC}_2 = \textsc{LIME}_2(Z, \textrm{uni})$
            \STATE $\textsc{MI}_2 = \textsc{LIME}_2(Z, \textrm{multi})$
        \ENDFOR
\end{algorithmic}
\end{algorithm}

\end{document}